\documentclass[10pt,twocolumn,letterpaper]{article}

\usepackage{iccv}
\usepackage{times}
\usepackage{epsfig}
\usepackage{graphicx}
\usepackage{amsmath}
\usepackage{amssymb}
\usepackage{multirow}
\usepackage{subfigure}
\usepackage{stfloats}
\usepackage[noend]{algpseudocode}
\usepackage{color}
\usepackage{mathtools}
\usepackage{booktabs}
\usepackage{hhline}
\usepackage{csquotes}
\usepackage{adjustbox}
\usepackage{diagbox}
\usepackage{arydshln}
\usepackage[ruled]{algorithm2e}
\def\eg{\emph{e.g.}} 
\def\ie{\emph{i.e.}} 

\usepackage[pagebackref=false,breaklinks=true,letterpaper=true,colorlinks,bookmarks=false]{hyperref}
\iccvfinalcopy 

\def\iccvPaperID{1072} 

\ificcvfinal\fi

\begin{document}
	
	\title{Weakly Supervised Person Search with Region Siamese Networks}
	\newcommand*\samethanks[1][\value{footnote}]{\footnotemark[#1]}
	\author{Chuchu Han$^1$\thanks{This work was done when Chuchu Han was an intern at ByteDance.}, Kai Su$^2$, Dongdong Yu$^2$, Zehuan Yuan$^2$, Changxin Gao$^1$,\\ Nong Sang$^1$\thanks{Corresponding author}, Yi Yang$^3$, Changhu Wang$^2$\\
	$^1$Key Laboratory of Ministry of Education for Image Processing and Intelligent Control, \\
	School of Artificial Intelligence and Automation, Huazhong University of Science and Technology\\
	$^2$ByteDance \quad $^3$University of Technology Sydney \\
{\tt\small {\{hcc, cgao, nsang\}@hust.edu.cn }} \quad {\tt\small {Yi.Yang@uts.edu.au}}\\
	{\tt\small {\{sukai, yudongdong, yuanzehuan, wangchanghu\}@bytedance.com}}   
	}
	
	\maketitle
	\ificcvfinal\thispagestyle{empty}\fi
	
\begin{abstract}
	Supervised learning is dominant in person search, but it requires elaborate labeling of bounding boxes and identities. Large-scale labeled training data is often difficult to collect, especially for person identities. A natural question is whether a good person search model can be trained without the need of identity supervision. In this paper, we present a weakly supervised setting where only bounding box annotations are available. Based on this new setting, we provide an effective baseline model termed Region Siamese Networks (R-SiamNets). Towards learning useful representations for recognition in the absence of identity labels, we supervise the R-SiamNet with instance-level consistency loss and cluster-level contrastive loss. For instance-level consistency learning, the R-SiamNet is constrained to extract consistent features from each person region with or without out-of-region context. For cluster-level contrastive learning, we enforce the aggregation of closest instances and the separation of dissimilar ones in feature space. Extensive experiments validate the utility of our weakly supervised method. Our model achieves the rank-1 of 87.1\% and mAP of 86.0\% on CUHK-SYSU benchmark, which surpasses several fully supervised methods, such as OIM~\cite{xiao2017joint} and MGTS~\cite{chen2018person}, by a clear margin. More promising performance can be reached by incorporating extra training data. We hope this work could encourage the future research in this field.
\end{abstract}
\vspace{-2.3mm}

\section{Introduction}

\begin{figure}[t]
	\centering
	\subfigure[Fully supervised setting]{%
		\label{fig:intro_1}%
		\includegraphics[width=0.95\linewidth]{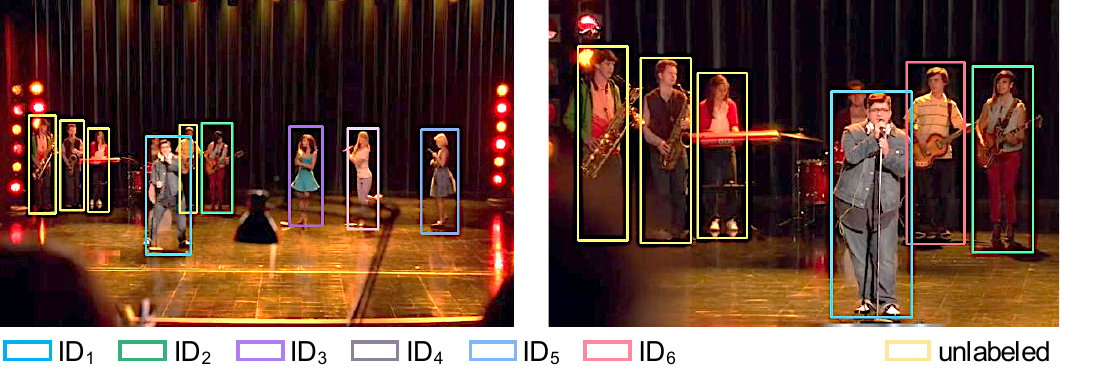}}%
	
	\subfigure[Proposed weakly supervised setting]{%
		\label{fig:intro_2}%
		\includegraphics[width=0.95\linewidth]{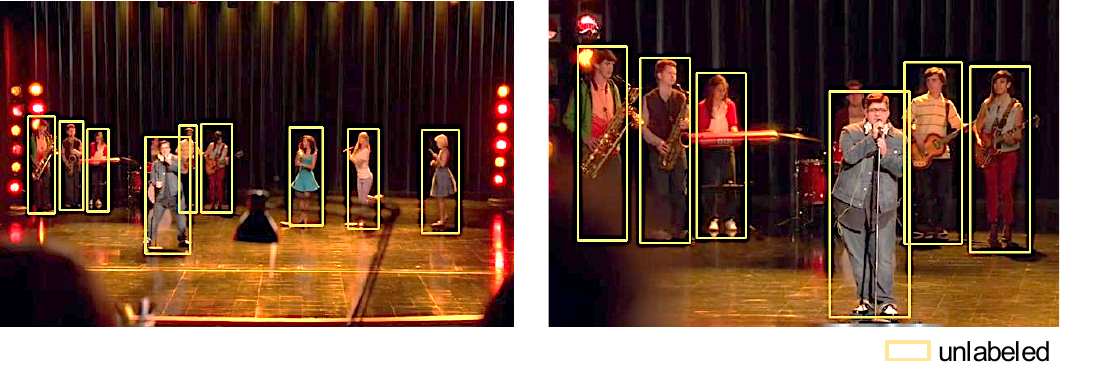}}%

	\caption{Comparisons between two settings. (a) Fully supervised setting. The images are annotated with both bounding boxes and person identities. Note that some identity annotations have lacked in original person search datasets. (b) The proposed weakly supervised setting. The images only have bounding box annotations.}
	\vspace{-3mm}
	\label{fig:intro}
\end{figure}

Person search~\cite{xiao2017joint} aims to localize and recognize a query person from a gallery of unconstrained scene images. Despite tremendous progress achieved by recent works~\cite{xiao2017joint,xiao2019ian,munjal2019query,yan2019learning,dong2020instance,dong2020bi,chen2020norm,zhong2020robust}, the training process requires strong supervision in terms of bounding boxes and identity labels, shown in Fig.~\ref{fig:intro_1}. However, obtaining such annotations at a large scale can be time-consuming and economically expensive. Even in the most widely used dataset, CUHK-SYSU~\cite{xiao2017joint}, almost 72.7\% of pedestrian bounding boxes have no identity annotations. It indicates that labeling identities is more difficult than bounding boxes. The infeasible identity annotations largely restrict the scalability of supervised methods.



Instead of relying on the expensive labeling, a lot of researchers have been dedicated to training models with no labels~\cite{he2020momentum,chen2020simple} or incomplete labels~\cite{oquab2015object,zhou2018brief} in the areas of image recognition, object detection, \etc. 
Nevertheless, relevant explorations are missing in the field of person search. To fill the gap, we investigate the person search modeling in a weakly supervised setting, where only bounding box annotations are required. As shown in Fig.~\ref{fig:intro_2}, the proposed setting alleviates the burden of obtaining manually labeled identities. 
However, it is more challenging to pursue accurate person search using sole bounding box annotations.

In this paper, we setup a strong weakly supervised baseline called Region Siamese Networks (R-SiamNets). Towards learning meaningful feature representations of each instance, our model minimizes the discrepancy between two encoded features transformed from the same pedestrian region. Specifically, one branch is fed with the whole scene image and extracts the RoI features of the person instance. The other branch extracts features from the cropped person image. The obtained features from the two weight-sharing branches are constrained to be consistent. The motivation of this design is that the context-free instance features extracted from the cropped image regions could help the network to distinguish persons from the irrelevant background content. We formulate a self-instance consistency loss and an inter-instance similarity consistency loss to supervise the learning of context-invariant feature representations. 

Further, a cluster-level contrastive learning method is introduced for striking a balance between separation and clustering. The clustering method aggregates the closest instances together and pushes instances from different clusters apart. It is assumed that the closest features have a high probability of being from the same individual. The generated pseudo labels by clustering are used for contrastive learning. We iteratively apply this non-parametric clustering for refinement along with the training process. The cluster-level contrastive learning yields a significant performance gain, with the $6.1\%$ absolute improvement of mAP on the CUHK-SYSU dataset. 

Our contributions can be summarized in three-folds:
\begin{itemize}
	\item We introduce a weakly-supervised setting for person search. The new setting only requires bounding box annotations, relieving the burden of obtaining manually labeled identities. 
	 With this setting, the developed algorithms could be readily utilized for large-scale person search in real-world scenarios.
	 
	\item We propose the R-SiamNet as a baseline under the weakly supervised setting. With the Siamese network, instance-level consistency learning is applied to encourage context-invariant representations. Further, cluster-level contrastive learning is introduced for striking a balance between separation and clustering.
	
	\item Our R-SiamNet achieves the rank-1 of 87.1\% and 75.2\% on CUHK-SYSU and PRW datasets, respectively. The results outperform several supervised methods by a clear margin, \eg, OIM~\cite{xiao2017joint} and MGTS~\cite{chen2018person}. More promisingly, the performance is further promoted when incorporating more extra datasets.
\end{itemize}

	\section{Related Work}
	\vspace{0.5em}
	\noindent\textbf{Person Search.}
	Recently, person search task has raised a lot of interest in the computer vision community~\cite{xiao2017joint,zheng2017person,chen2018person,chang2018rcaa,lan2018person,chen2020norm,zhang2021diverse}.
	In the literature, there are two manners to deal with this problem, \ie, two-step and one-step. 
	
	For two-step methods, the pedestrian detection and person re-identification are trained with two separated models~\cite{zheng2017person,chen2018person,chang2018rcaa,lan2018person,han2019re}. Zheng~\textit{et~al.}~\cite{zheng2017person} evaluate various combinations of different detectors and re-ID networks, and develop a Confidence Weighted Similarity (CWS) to assist the pedestrian matching. Chen~\textit{et~al.}~\cite{chen2018person} enhance the feature representations by introducing a mask-guided two-steam model. Han~\textit{et~al.}~\cite{han2019re} develop an RoI transform layer to optimize the two networks end-to-end. With more parameters, these methods guarantee high performance while low efficiency in evaluation. 
	
	One-step methods~\cite{xiao2017joint,xiao2019ian,han2021decoupled,chen2020norm} jointly train the detection and re-ID in a unified model, exhibiting high efficiency. Among these methods,~\cite{xiao2017joint, xiao2019ian} take the Faster R-CNN~\cite{ren2015faster} as a backbone network, and most layers are shared by two tasks. Munjal~\textit{et~al.}~\cite{munjal2019query} first introduce a query-guided end-to-end person search network. With the global context from both query and gallery images, the well-designed framework generates query-relevant proposals and learns query-guided re-ID scores. Yan~\textit{et~al.}~\cite{yan2019learning} explore the contextual information and build a graph learning framework to employ context pairs to update target similarity. Dong~\textit{et~al.}~\cite{dong2020bi} develop a bi-directional interaction network and employ the cropped person patches as the guidance to reduce redundant context influences. 
	
	These studies are fully supervised and require precise annotations for each person, including the bounding boxes and the identities. It is impractical to extend these methods in large-scale scenarios due to the expensive labeling process. Thus, we introduce a weakly supervised setting and develop a framework trained solely with bounding boxes.

	\vspace{0.5em}
	\noindent\textbf{Siamese Networks.}
	Siamese networks~\cite{bromley1994signature} consist of twin networks which accept distinct inputs, and the comparability is determined by supervision. This architecture is widely used in many fields, including object tracking~\cite{bertinetto2016fully}, one-shot learning~\cite{koch2015siamese}, signature~\cite{bromley1994signature} and face verification~\cite{taigman2014deepface}, \etc. In this paper, we explore the context-invariant embeddings based on a region Siamese network with two forms of inputs, \ie, whole scene images, and cropped images.

	\vspace{0.5em}
	\noindent\textbf{Contrastive Learning.}
	Contrastive learning~\cite{hadsell2006dimensionality} aims at attracting the positive sample pairs and repulsing the negative ones, which has been popularized in recent unsupervised learning~\cite{hjelm2018learning, wu2018unsupervised, ye2019unsupervised, he2020momentum, chen2020simple, chen2020exploring}. Wu~\textit{et~al.}~\cite{wu2018unsupervised} consider each instance as a class, and employ a memory bank to store the instance embeddings. Similar to~\cite{wu2018unsupervised}, Ye~\textit{et~al.}~\cite{ye2019unsupervised} learn the data augmentation invariant and instance spread-out features. 
	MoCo~\cite{he2020momentum} maintains the dictionary with a queue and a moving-averaged encoder, so the contrastive learning is viewed as the dictionary look-up. SimCLR~\cite{chen2020simple} deprecates the usage of the memory bank, and directly uses negative samples in the current batch. 
	In this paper, rather than only independently penalizing the incompatibility of each single positive pair at a time, we construct more informative positive pairs by non-parametric clustering. 


	\vspace{0.5em}
	\noindent\textbf{Unsupervised Person Re-identification.}
	The traditional unsupervised methods can be summarized into three categories, designing hand-craft features, utilizing localized salience statistics, or dictionary learning-based works. However, the performance of these methods is inferior to supervised ones. On the basis of clustering algorithms, recent works exhibit higher performance. Lin~\textit{et~al.}~\cite{lin2019bottom} develop a bottom-up clustering framework to iteratively train the network with pseudo labels. Zeng~\textit{et~al.}~\cite{zeng2020hierarchical} combine the hierarchical clustering method with hard-batch triplet loss. Ge~\textit{et~al.}~\cite{ge2020self} design a self-paced contrastive learning strategy with a novel clustering reliability criterion to filer the unstable clusters with DBSCAN~\cite{ester1996density}. Although the cluster-based methods achieve high performance, they require carefully adjusted hyperparameters, \eg, number of clusters~\cite{lin2019bottom, zeng2020hierarchical}, or the distance threshold~\cite{ester1996density}. 
	
	In this paper, we employ a non-parametric clustering method~\cite{sarfraz2019efficient}, with a filtering mechanism by image information. This manner only relies on the first neighbor of each data point, and requires no hyper-parameters.


	\section{Proposed Method}
	In this section, we first introduce the overall region Siamese network in Sec.~\ref{sec:3a}. Then we will describe the instance-level consistency learning in Sec.~\ref{sec:3b}.
	At last, the cluster-level contrastive learning is developed in Sec.~\ref{sec:3c}.
	
	\subsection{Region Siamese Network}  \label{sec:3a}
	Our main target is to locate the person positions and learn representative features for identification. Under the weakly supervised setting, only the bounding box annotations are employed in the training process. Without manually labeled identities, it is essential to design supervision signals for training the network. To achieve this goal, we develop the framework from two aspects: 
	1) Based on Siamese networks, the comparability is determined by supervision with different augmented inputs. In this paper, both whole scene images and cropped images are taken as inputs. We focus on the instance-level consistency learning to encourage context-invariant embeddings.
	2) Pseudo labels generated by clustering permit to model cluster-level supervision. Thus, we apply the cluster-level contrastive learning, reaching a balance between separation and aggregation.

\begin{figure*}[htbp]
		\small
		\begin{center}
			\includegraphics[width=0.9\linewidth]{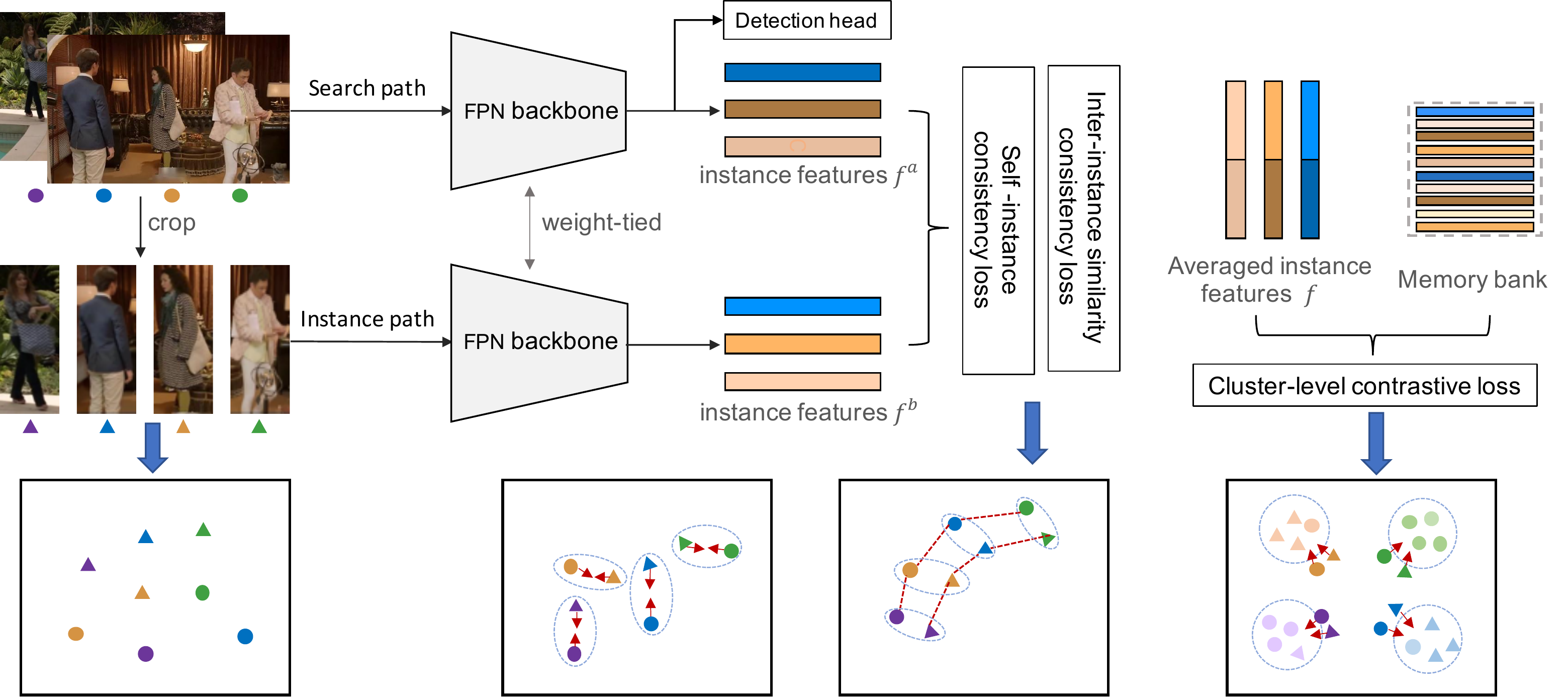} 
		\end{center}
		\caption{Illustration of our R-SiamNet for weakly supervised person search. Given whole scene images, the detection and identification are trained jointly with the backbone in the search path. The features of pedestrians are produced by the RoI align layer, denoted as $f^a$. Meanwhile, we introduce an instance path with the cropped persons as inputs. In this path, we extract the features $f^b$ through the same layers. To encourage the context-invariant features, a self-instance consistency loss and an inter-instance similarity consistency loss are applied between $f^a$ and $f^b$. Besides, pseudo labels are produced with the non-parametric clustering. We calculate the cluster-level contrastive loss between the averaged features $f$ and the embeddings in memory bank. Note that only the search path is utilized in testing.}
		\label{fig:network}
	\end{figure*}

	Based on these considerations, we propose a region Siamese network as shown in Fig.~\ref{fig:network}. There are two branches termed as search path and instance path. With an extra detection head, the search path takes the whole scene images as inputs, training the detection and identification jointly. The feature embedding of each pedestrian is generated by the Region-of-Interest (RoI) align layer. For the instance path, the cropped images are taken as inputs. With less context, the corresponding outputs can focus on the regions of pedestrians.
	To ensure the context-invariant embeddings, we apply the instance-level consistency learning on the output of two paths. It consists of a self-instance consistency loss and an inter-instance similarity consistency loss. Besides, to model the cluster-level supervision, we employ a non-parametric clustering method based on the nearest neighbor of each sample. A cluster-level contrastive loss is developed to calculate the similarities between the samples in the current batch and the memory bank. During inference, we only utilize the search path.
	
	
	
	\subsection{Instance-Level Consistency Learning}  \label{sec:3b}
	With no identity annotations, it is observed that the learned instance features are involved with excessive context. As a fine-grained task, the retrieval process is easily affected by interference from surrounding persons and noises. To alleviate this issue, we input the cropped pedestrians which contain less context to build the supervision. The instance-level consistency learning is developed to encourage the context-invariant embeddings from two aspects.
	
	\vspace{0.5em}
	\noindent\textbf{Self-instance consistency loss.} Given a mini-batch of scene images, we obtain $B$ cropped pedestrian images with bounding box labels. The scene images and cropped ones are processed by the region Siamese network. The output embeddings are denoted as $\textbf{F}_a = [f_1^a, f_2^a, ..., f_B^a]^T$ and $\textbf{F}_b = [f_1^b, f_2^b, ..., f_B^b]^T$ for search path and instance path, respectively. 
	To encourage the context-invariant embeddings for a specific instance, we consider to maximize the cosine similarity between $f^a$ and $f^b$. Then the self-instance consistency loss is defined as:
	\begin{equation}
		\begin{aligned}
			L_{ins} = \frac{1}{B} \sum_{i=1}^{B} (1 - \frac{f^a_i}{||f^a_i||_2} \cdot \frac{ f^b_i}{ ||f^b_i||_2}),
		\end{aligned}
		\label{eq:L_in}
	\end{equation}
	where $||\cdot||_2$ is the L2 norm, and the loss is averaged over all instances in a mini-batch.

	\vspace{0.5em}
	\noindent\textbf{Inter-instance similarity consistency loss.} 
	The feature embeddings in each path can be seen as an aggregated feature space. The above self-instance consistency loss only constrains the embedding pairs from the same person to be closer individually. We further apply a constraint to enlarge the similarity distribution between the entire feature spaces. For the search path, the similarity matrix $\textbf{S}^a \in {\mathbb{R}^{B \times B}}$ is obtained by $\textbf{S}^a =\overline{\textbf{F}}_a^{}   \overline{\textbf{F}}_a^{T}$, where $\overline{\textbf{F}}_a^{}$ is produced by a row-wise L2 normalization on $\textbf{F}_a^{}$. In the same way, $\textbf{S}^b  \in {\mathbb{R}^{B \times B}}$ is calculated for the instance path. Our goal is to keep the consistency between the two similarity distributions. Based on the  divergence, we develop an inter-instance similarity consistency loss as follows:
	\begin{equation}
		\begin{aligned}
			L_{int} = D_{KL}(\textbf{S}^a || \textbf{S}^b) + D_{KL}(\textbf{S}^b || \textbf{S}^a),
		\end{aligned}
		\label{eq:L_kl}
	\end{equation}
	where $D_{KL}$ denotes the KL divergence. It can be formulated as $ D_{KL}(P || Q)=\sum_{x \in {\mathbb{X}}}P(x)log({P(x)}/{Q(x)})$. Since each distribution can be considered as the target, we adopt the mutual manner.

	\subsection{Cluster-Level Contrastive Learning}  \label{sec:3c}
	The instance recognition~\cite{wu2018unsupervised} treats each sample as a category to produce well-separated samples. In person search task, this approach may be inferior. We aim at exploring the similarities among persons and striking a balance between separation and aggregation. A non-parametric clustering method is employed to produce pseudo labels, and a cluster-level contrastive loss is applied between samples in the current batch and the memory bank.
	
	\vspace{0.5em}
	\noindent\textbf{Non-parametric clustering.}
	We build the cluster-level supervision based on an assumption that the most similar feature embeddings have high probabilities of belonging to the same class. This motivates our clustering manner, in which only the nearest neighbor of each sample is aggregated. Meanwhile, there is a prior that the pedestrians in same scene images belong to different identities. Thus, we can filter some false aggregations when clustering. After clustering, all persons are assigned pseudo labels.
	 
	Specifically, at the beginning of each epoch, the clustering process is conducted after extracting the embeddings of all instances. Supposing there are $N$ samples, an adjacency matrix $\textbf{A}(i,j) \in \mathbb{R}^{N \times N}$ can be constructed among all samples, which is initialized with all zeros. To set $\textbf{A}(i,j)=1$, two conditions should be satisfied simultaneously: 
	
	1) $j= \kappa_i^1$ or $\kappa_j^1=i$ or $\kappa_i^1=\kappa_j^1$, where $\kappa_i^1$ denotes the first neighbor of sample $i$. Besides the nearest neighbors, the adjacency matrix also links points that have the same neighbor with $\kappa_i^1=\kappa_j^1$. 
	
	2) The whole scene image of the $i$-th and $j$-th pedestrian should be different. If two persons come from the same scene image, they cannot be clustered.
	

	
	Different from other clustering algorithms that require carefully adjusted hyperparameters~\cite{lin2019bottom,ge2020self,zeng2020hierarchical}, \eg, cluster numbers or distance threshold, we employ the non-parametric clustering approach. It is easily scalable to large-scale data with minimal computational expenses.

	\vspace{0.5em}
	\noindent\textbf{Cluster-level contrastive loss.}
	Similar to previous works~\cite{xiao2017joint,chen2020norm}, we maintain a memory bank $\textbf{M} \in \mathbb{R}^{N \times d}$ to store the embeddings of all instances, where $d$ denotes the feature dimensions. After clustering, the memory bank is updated by the newly embedded features with pseudo labels. Then, a cluster-level contrastive loss can be computed between the features in the memory bank and current batch.
	
	In a mini-batch, a specific instance feature is denoted as $f = mean(f^a, f^b)$. Assuming that there are $K$ positive samples in the memory bank sharing the same pseudo label with $f$. Then, the remaining $J$ samples in $\textbf{M}$ are considered as negative samples. The cosine similarities are denoted as $\{s_p^i\} (i=1,2,...,K)$ and $\{ s_n^j\} (j=1,2,...,J)$, respectively. Inspired by~\cite{sun2020circle}, we apply the cluster-level contrastive loss to make each $s_p^i$ greater than $s_n^j$:
	\begin{equation}
		\begin{aligned}
			L_{clu} = log[1+ \sum_{i=1}^{K} \sum_{j=1}^{J} \exp(\gamma(s_n^j-s_p^i))],\\
		\end{aligned}
		\label{eq:L_cl}
	\end{equation}
	where $\gamma$ is the scale factor. During backward, the memory bank is updated with the samples in current mini-batch: $M_t \leftarrow \lambda M_t + (1-\lambda) f$. $\lambda$ is the momentum factor and $t$ denotes the instance position in the memory bank.

	\begin{algorithm}[t]
		\small
		\SetAlgoLined
		\SetKwInOut{Input}{Input}
		\SetKwInOut{Output}{Output}
		\SetKwInput{Initialization}{Initialization}
		\caption{Training procedure of R-SiamNet}\label{alg:weakly-supervised}
		\Input{Unlabeled data $\textbf{I} = \{I_1, I_2, ..., I_{N'} \}$; \\
			Scale factor $\gamma$; Momentum $\lambda$}
		\Initialization{Initialize the backbone with ImageNet-pretrained ResNet-50.} 
		\For{each epoch} {
			\textbf{1:} Extract all the instance features.\\
			\textbf{2:} Conduct the non-parametric clustering. \\
			\textbf{3:} Update the features in the memory bank.\\
			\For{each mini-batch} {
				\textbf{1:} Encode instance features of two paths through R-SiamNet: \\
				\qquad $f^a = \Phi_{\theta} (\textbf{I})$ \\
				\qquad $f^b = \Phi_{\theta} (crop(\textbf{I}))$\\
				\qquad $f = mean(f^a, f^b)$\\
				\textbf{2:} Compute the self-instance consistency loss with Eq.\ref{eq:L_in}\\
				\textbf{3:} Compute the inter-instance similarity consistency loss with Eq.~\ref{eq:L_kl}\\
				\textbf{4:} Compute the cluster-level contrastive loss with Eq.\ref{eq:L_cl}\\
				\textbf{5:} During backward, update the features in memory bank: \\
				\qquad $M_t \leftarrow \lambda M_t + (1-\lambda) f$			
			}
		}
	\end{algorithm}

	\subsection{Training Procedure}  \label{sec:3d}
	Given the input images, we aim to learn a deep convolutional neural network (CNN) model $\Phi_{\theta}$ to realize precise localization and identification. The details of our training procedure are provided in Algorithm~\ref{alg:weakly-supervised}. 
	In summary, our total training objective is revised as:
	\begin{equation}
		\begin{aligned}
			L = L_{ins} + L_{int} + L_{clu} + L_{det}, \\
		\end{aligned}
		\label{eq:L_all}
	\end{equation}
	where $L_{det}$ denotes the detection losses, including regression loss and foreground-background classification loss.

	\section{Experiments}
	In this section, we first introduce two benchmark datasets, followed by the settings under weakly supervised manner and evaluation protocols in Sec.~\ref{sec:4a}. Then Sec.~\ref{sec:4b} describes the implementation details and the reproducibility. We conduct a series of ablation studies to analyze the effectiveness of the proposed method in Sec.~\ref{sec:4c}. Finally, we discuss the comparison with the state of the arts in Sec.~\ref{sec:4d}.

	\subsection{Datasets and Settings} \label{sec:4a}
	\vspace{0.5em}
	\noindent\textbf{CUHK-SYSU dataset.} 
	CUHK-SYSU~\cite{xiao2017joint} is a large-scale person search dataset, which is composed of urban scene pictures and movie snapshots. There are $18,184$ images with $96,143$ annotated bounding boxes, including $8,432$ labeled identities. The training set consists of $11,206$ images with $5,532$ identities and several unlabeled ones. There are $6,978$ gallery images and $2,900$ probe images in testing set.
	
	\vspace{0.5em}
	\noindent\textbf{PRW dataset.} 
	PRW~\cite{zheng2017person} is captured by six spatially disjoint cameras in the university. It consists of $11,816$ frames with $43,110$ annotated bounding boxes, among which $34,304$ are assigned with $932$ identity labels, and the rest are unlabeled ones. The training set contains $5,704$ frames with $482$ identities, and the testing set includes $6,112$ gallery images and $2,057$ queries with $450$ identities.
	
	\vspace{0.5em}
	\noindent\textbf{Settings.}
	Under the fully supervised setting, the statistics of the training data are shown in Tab.~\ref{tab:statistics}. In this paper, we propose a weakly supervised setting for person search, reducing the need of strong supervision during training. Under the weakly supervised setting, our model is trained only using $55,260$ and $18,048$ bounding box annotations for the CUHK-SYSU and PRW datasets, respectively.
	
	\begin{table}[b]
		\footnotesize
		\begin{center}
			\caption{Training data statistics on the CUHK-SYSU and PRW datasets within the fully supervised settings. Bbox: Bounding box.}
			\label{tab:statistics}
			\begin{tabular}{ccccc}
				\toprule
				\multirow{2}*{Dataset} &
				\multirow{2}*{Images} &
				\multirow{2}*{IDs} &
				\multicolumn{2}{c}{Bboxes} \\
				\cmidrule{4-5}
				~&~&~&Labeled&Unlabeled\\		
				\midrule	
				CUHK-SYSU &11206 & 5532 & 15080 (27.3\%) & 40180 (72.7\%)   \\			
				PRW& 5704 & 482 & 14906 (82.6\%) & 3142 (17.4\%) \\
				\bottomrule
			\end{tabular}
		\end{center}
	\end{table}
	
	 
	\vspace{0.5em}
	\noindent\textbf{Evaluation protocols.} 
	Our experiments employ the standard evaluation metrics in person search~\cite{xiao2017joint}. One is the cumulative matching cure (CMC), which is inherited from the person re-ID. A candidate is counted if the intersection-over-union (IoU) with ground truth is greater than 0.5. The other is the mean Average Precision (mAP), and it is inspired by the object detection task. For each query, we compute an averaged precision (AP) based on the precision-recall curve. Then, the mAP is calculated by averaging the APs across all the queries.

	\subsection{Implementation Details}  \label{sec:4b}
	\vspace{0.5em}
	\noindent\textbf{Model.} 
	We employ RepPoints~\cite{yang2019reppoints} released by OpenMMLab~\cite{mmdetection} as our backbone network, containing the ImageNet pretrained ResNet-50~\cite{he2016deep}, the feature pyramid network (FPN)~\cite{lin2017feature}, and a detection head. The search path takes the scene images as inputs, jointly training the detection and recognition. To obtain the pedestrian features $f^a$, the RoI align is applied on FPN with the ground truth RoIs, followed by a fully connected (fc) layer after flattening. For the instance path, the cropped and resized images are taken as inputs. Similar to $f^a$, $f^b$ is produced with the same network except the RoI align, and both features are $2048$-d.

	\vspace{0.5em}
	\noindent\textbf{Training.} 
	The scene images are resized to $1333 \times 800$, and cropped images are rescaled to $192 \times 64$. The batched Stochastic Gradient Descent (SGD) optimizer is used with a momentum of $0.9$. The weight decay factor for L2 regularization is set to $5\times10^{-4}$. We use a mini-batch size of $9$, and an initial learning rate of $1\times10^{-3}$. The model is trained for $48$ epochs with the learning rate multiplied by $0.1$ at $32$ and $44$ epochs. The scale factor $\gamma$ is set to $16$ and the momentum $\lambda$ is set to $0.2$ for both datasets. All experiments are implemented on the PyTorch framework, and the network is trained on the NVIDIA Tesla V100.
	
	\subsection{Ablation Study}  \label{sec:4c}
	To evaluate the effectiveness of the proposed framework, we conduct detailed ablation studies on the CUHK-SYSU and PRW datasets. Note that all the settings in each experiment are the same as the implementation in Sec.~\ref{sec:4b}.

	\vspace{0.5em}
	\noindent\textbf{Effectiveness of different components.} 
	To verify the effectiveness of each component, we compare the performance under different settings in the training process. The results are shown in Tab.~\ref{tab:component analysis}. 

	\textit{Instance Recognition (IR)}~\cite{wu2018unsupervised} denotes that each instance is treated as a category in training. A memory bank is maintained to store all the instance features, providing abundant negative samples to compute the contrastive loss. 
	\textit{Search path w/ IR} means the supervision of IR is only applied on the instance features produced by the search path. This approach can be viewed as the baseline of our method. With the scene images as input, instance features may contain excessive context to disturb the matching, thus the mAP only reaches $51.85\%$. 
	\textit{Instance path w/ IR} indicates the IR is applied on the features generated by the instance path, which takes the cropped persons as inputs. In the search path, only the detection head is trained within the same backbone. Under this setting, the result can reach $63.79\%$ on mAP. Compared to the search path, the instance path contains less context, exhibiting higher performance. 
	%
	
	 \textit{R-SiamNet w/ $L_{ins}$} takes both scene images and cropped pedestrian images as inputs, and the fused outputs are supervised by IR. To encourage the context-invariant feature embeddings, we apply the \textit{self-instance consistency loss $L_{ins}$} between two paths. It maximizes the similarity of pair-wise features from two paths. The mAP is promoted to $76.06\%$, surpassing $51.85\%$ by a large margin. This verifies the importance of keeping consistency. For further restraint, we develop the \textit{inter-instance similarity consistency loss $L_{int}$}, which is applied on the similarity distributions within the mini-batch of two paths. This further guarantees the context-invariant embeddings, and achieves a gain of $3.96\%$ on rank-1.
	Moreover, to explore the cluster-level supervision, we implement the non-parametric clustering to produce pseudo labels. Thus, a \textit{cluster-level contrastive loss $L_{clu}$} is employed for supervision instead of IR. From Tab.~\ref{tab:component analysis}, we can see that the performance achieves  $85.72\%$ on mAP and $86.86\%$ on rank-1. The results show the effectiveness of our framework.
	%
	%
	\begin{table}[htbp]
		\footnotesize
		\centering
		\caption{Component analysis of our method. IR: Instance recognition. The rank-1/5/10 accuracy (\%) and mAP (\%) are shown.}
		\vspace{1mm}
		\label{tab:component analysis}
		\setlength{\tabcolsep}{1.7mm}{
			\begin{tabular}{lcccc}
				\toprule	
				\multirow{2}{*}{Method} &
				\multicolumn{4}{c}{CUHK-SYSU} \\
				\cmidrule{2-5}
				~&mAP&R1&R5&R10 \\
				\midrule	
				\midrule				
				Search path w/ IR&51.85&59.69&67.31&69.03\\	
				Instance path w/ IR&63.79&65.55&82.21&86.83\\	
				R-SiamNet w/  $L_{ins}$ \&  IR&76.06&78.21&90.28&92.90\\	
				R-SiamNet w/  $L_{ins}$ \& $L_{int}$ \&  IR&79.62&82.17&91.69&94.03\\	
				R-SiamNet w/ $L_{ins}$ \& $L_{int}$ \& $L_{clu}$&85.72&86.86&95.24&96.86\\		
				\bottomrule	
			\end{tabular}
		}
	\end{table}

	\begin{figure*}[htbp]
		\small
		\begin{center}
			\includegraphics[width=0.95\linewidth]{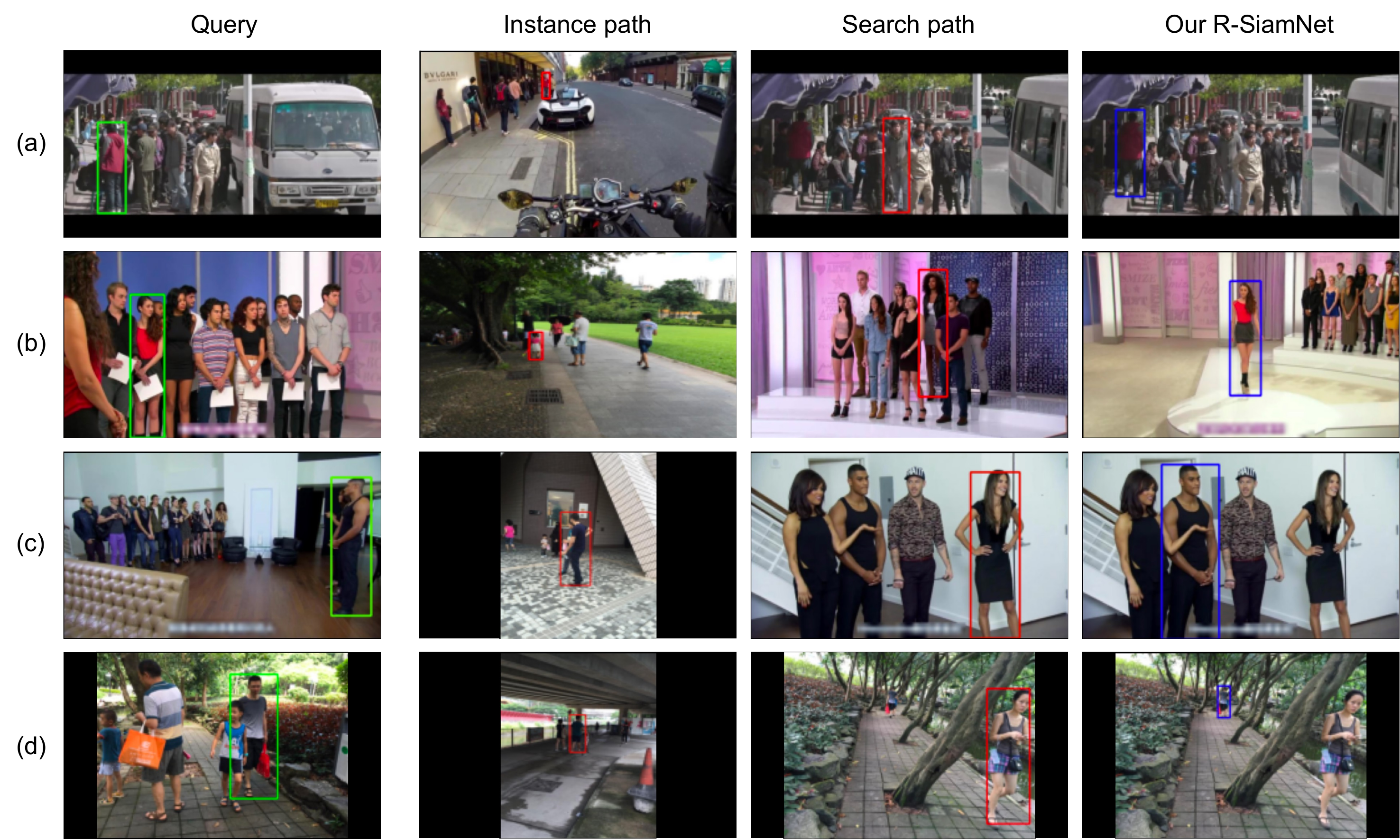}
		\end{center}
		\caption{Visualization of different methods on the CUHK-SYSU dataset. Given the query images, we show the rank-1 search results of three training approaches. First column shows the query persons with the green boxes. The instance path/search path denote the model is trained with a single path. The last column show the result of our region Siamese network. Red/blue boxes represent the wrong/correct results, respectively.}
		\label{fig:vis}
	\end{figure*}
	
	\vspace{0.5em}
	\noindent\textbf{Scalability with different scales of training data.} 
	Our framework is designed to learn discriminative identity embeddings under the weakly supervised setting. The scalability is essential for the methods when giving more training data without identity labeling. To discuss the scalability, we design the experiment from two aspects.

	First, we assess the performance with different percentages of training data individually, as shown above the dotted line in Tab.~\ref{tab:data_size}. Specifically, we divide the CUHK-SYSU/PRW dataset by 20\%, 40\%, 60\%, 80\%, 100\% for training. It can be seen that the performance is gradually improved as the training scale increases. Moreover, the growing tendency has not reached saturation on both datasets, indicating that the proposed framework can achieve further improvement with more training data. 
	
	
	Second, to further evaluate the scalability of our method, we expand the training set by combining different datasets. The results are shown below the dotted line in Tab.~\ref{tab:data_size}. When training with CUHK-SYSU and PRW datasets together, the performance on both datasets has been considerably improved. Especially for the PRW dataset, the mAP is promoted by a large margin since the added CUHK-SYSU owns a larger data scale. Besides, we also employ a dataset called INRIA~\cite{dalal2005histograms} in the pedestrian detection task. There are $902$ images that contain $1,826$ pedestrians with bounding boxes labeling. When training with the three datasets together, our performance is further promoted on both CUHK-SYSU and PRW datasets. 
	
	All the experiments prove that our framework has the potential to reach promising performance by incorporating more training data.

	\begin{table}[htbp]
		\footnotesize
		\centering
		\caption{Performance with different scales of training set. Above the dotted line, the results with different percentages of training data are shown. Below the dotted line, it exhibits the performance of combining with more training datasets.}
		\vspace{1mm}
		\label{tab:data_size}
		\begin{tabular}{ccccc}
			\toprule	
			\multirow{2}{*}{Training set} &
			\multicolumn{2}{c}{CUHK-SYSU} &
			\multicolumn{2}{c}{PRW} \\
			\cmidrule{2-5}
			~&R1&mAP&R1&mAP \\
			\midrule	
			\midrule				
			20\% data &78.34 & 76.71&66.94 &15.61\\
			40\% data   &82.41 &80.50 &69.32& 17.59\\
			60\% data   &83.45  &82.52 &71.80& 19.10 \\
			80\% data   &85.41  &84.15&72.73 & 19.64\\
			100\% data   &86.86 &85.72 &73.36& 21.16 \\
			\hdashline
			CUHK-SYSU \&  PRW &87.00 & 85.92&75.06 & 23.50\\	
			CUHK-SYSU \&  PRW  \& INRIA &\textbf{87.59} & \textbf{86.19}&\textbf{76.03}& \textbf{25.53}\\	
			\bottomrule	
		\end{tabular}
	\end{table}
	
	\vspace{0.5em}
	\noindent\textbf{Visualization and Analysis.}
	To evaluate the effectiveness of the proposed method, we illustrate some qualitative search results on the CHUK-SYSU dataset. As Fig.~\ref{fig:vis} shows, we present the comparisons on rank-1 of our R-SiamNet and the other two manners. Specifically, the first column shows the query persons with green bounding boxes. Instance path/search path represents that the model employs the cropped images/whole scene images as input within a single path. The last column is the result of our R-SiamNet. The search results are exhibited with different colors, \ie, red boxes denote the wrong matches while the blue boxes show the correct ones. 
	
	There are several observations from the visualization. 
	First, it is observed that the wrong matches in the search path are definitely different from the query, but with similar contexts. This verifies the existence of excessive contexts, which disturb the matching process by involving more surrounding persons/backgrounds.
	Second, we find that most false examples in the instance path have a similar appearance with the query. This indicates the features are hardly disturbed by the contexts.
	Third, our R-SiamNet promotes the complementary of both paths, maintaining useful context to aid the person search. The results also show the effectiveness of our method, which can successfully localize and match the query person in most cases.

	\subsection{Comparisons with the State-of-the-arts} \label{sec:4d}
	In this section, we compare our proposed framework with current state-of-the-art methods on person search in Tab.~\ref{tab:sota}.
	The results of two-step methods~\cite{chang2018rcaa,chen2018person,lan2018person,han2019re,wang2020tcts} are shown in the upper block while the one-step methods~\cite{xiao2017joint,xiao2019ian,liu2017neural,yan2019learning,zhang2020tasks,munjal2019query,chen2020norm} in the lower block.

	\begin{table}[t]
		\small
		\begin{center}
			\caption{Experimental comparisons with state-of-the-art methods on the CUHK-SYSU and PRW datasets.}
			\label{tab:sota}
			\vspace{1mm}
			\begin{tabular}{l|lcccc}
				\toprule
				~&\multirow{2}*{Methods} & \multicolumn{2}{c}{CUHK-SYSU} & \multicolumn{2}{c}{PRW} \\
				\cline{3-4}
				\cline{5-6}
				~&~&R1&mAP&R1&mAP\\
				\midrule			
				\midrule
				~&\multicolumn{5}{l}{\textbf{\textit{Fully supervised setting:}}}  \\
				\multirow{5}*{\rotatebox{90}{two-step}}
				&RCAA~\cite{chang2018rcaa} & 81.3 & 79.3&-&- \\
				~&MGTS~\cite{chen2018person}   & 83.7& 83.0& 72.1 & 32.6  \\
				~&CLSA~\cite{lan2018person}   & 88.5 & 87.2& 65.0 & 38.7 \\
				~&RDLR~\cite{han2019re} & 94.2& 93.0& 70.2& 42.9    \\
				~&TCTS~\cite{wang2020tcts} & \textbf{95.1}& \textbf{93.9}& \textbf{87.5} & \textbf{46.8}  \\
				\midrule
				\multirow{12}*{\rotatebox{90}{one-step}}
				~&OIM~\cite{xiao2017joint}    & 78.7& 75.5  & 49.9& 21.3\\
				~&IAN~\cite{xiao2019ian}   & 80.1 & 76.3& 61.9 & 23.0\\
				~&NPSM~\cite{liu2017neural}  & 81.2& 77.9 & 53.1 &24.2  \\
				~&CTXGraph~\cite{yan2019learning}  & 86.5& 84.1 & 73.6 &  33.4\\
				~&DC-I-Net~\cite{zhang2020tasks}  & 86.5  & 86.2 & 55.1  & 31.8\\
				~&QEEPS~\cite{munjal2019query}  & 89.1& 88.9& 76.7 & 37.1  \\
				~&NAE~\cite{chen2020norm}  & 92.4& 91.5& 80.9& 43.3 \\
				~&NAE+~\cite{chen2020norm} & 92.9 & 92.1 & 81.1 & 44.0\\
				~&DMRNet~\cite{han2021decoupled} & 94.2 & 93.2  & 83.3 & 46.9 \\
				\cmidrule(lr){2-6}
				~&\multicolumn{5}{l}{\textbf{\textit{Weakly supervised setting:}}}  \\
				~&Ours (1333*800)  & \textbf{86.9} &\textbf{85.7}&\textbf{73.4} & \textbf{21.2}  \\
				~&Ours (1500*900)  & \textbf{87.1} &\textbf{86.0} &\textbf{75.2} & \textbf{21.4}\\
				\bottomrule
			\end{tabular}
		\end{center}
	\end{table}

	\vspace{0.5em}
	\noindent\textbf{Evaluation on CUHK-SYSU.} 
	The comparisons between our network and existing supervised methods on the CUHK-SYSU dataset are shown in Tab.~\ref{tab:sota}. When the gallery size is set to 100, our proposed method reaches $85.7\%$ on mAP and $86.9\%$ on rank-1. The performance can outperform several supervised methods. Moreover, when using a larger resolution of $1500 \times 900$, our results are further improved and reach $86.0\%$ on mAP and $87.1\%$ on rank-1.

\begin{figure}[t]
		\begin{center}
			\includegraphics[width=0.95\linewidth]{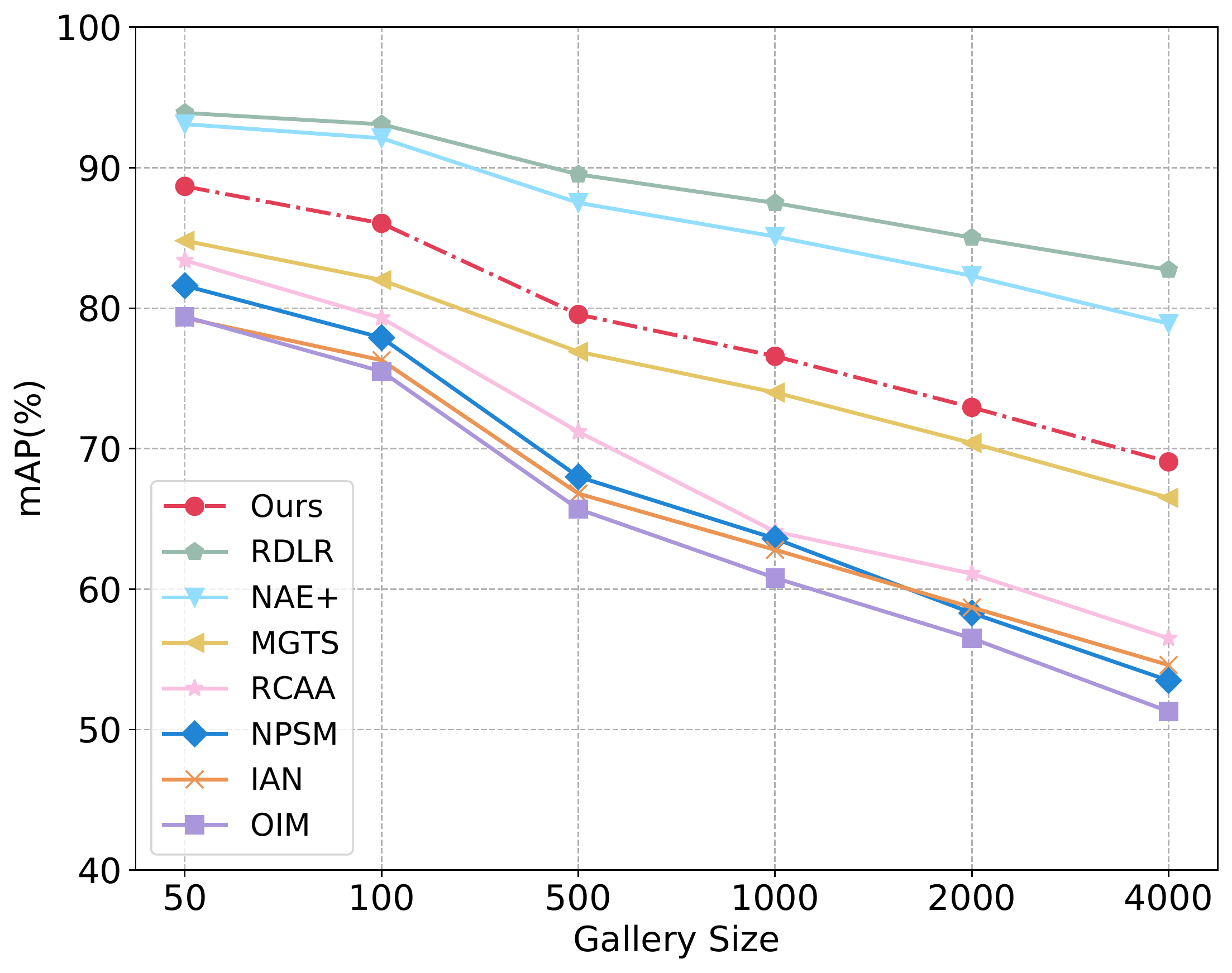}
		\end{center}
		\caption{Comparisons with different gallery sizes on the CUHK-SYSU dataset.
			Our method is represented by dotted lines.}
		\label{fig:gallery_size}
	\end{figure}

	To evaluate the performance consistency, we also compare with other competitive methods under varying gallery sizes of $[50, 100, 500, 1000, 2000, 4000]$. Fig.~\ref{fig:gallery_size} shows the comparisons with both one-step and two-step methods.
	It can be seen that the performance of all methods decreases as the gallery size increases. This indicates it is challenging when more distracting people are involved in the identity matching process, which is close to real-world applications. 
	Our method outperforms some supervised methods under different gallery sizes.

	\vspace{0.5em}
	\noindent\textbf{Evaluation on PRW.} 
	We also evaluate our method on the PRW dataset, as shown in Tab.~\ref{tab:sota}. Following the setting in benchmark~\cite{zheng2017person}, the gallery contains all the $6,112$ testing images. This is challenging since a tremendous number of detected bounding boxes are involved. Compared with the competitive techniques, our performance achieves $73.4\%$ on rank-1, and it is further promoted with larger resolutions.  Our method surpasses most works in both one-step and two-step manners. However, the results exhibit a low mAP due to the minor inter-class variations in this dataset.

	\section{Conclusion}
	In this paper, we introduce a weakly supervised setting for the person search task to alleviate the burden of costly labeling. Under this new setting, no specific identity annotations of pedestrians are required, and we only utilize the accessible bounding boxes for training. Meanwhile, we propose a baseline called R-SiamNet for localizing persons and learning discriminative feature representations. 
	To encourage the context-invariant features, a self-instance consistency loss and an inter-instance similarity consistency loss are developed.
	We also explore the balance between separation and aggregation by a cluster-level contrastive loss. Extensive experimental results on the widely used benchmarks demonstrate the effectiveness of our framework. The results also show that the gap with supervised state-of-the-arts will be further narrowed with more training data.
	
	\section*{Acknowledgements}	
	This work was partially supported by the Project of the National Natural Science Foundation of China No. 61876210, the Fundamental Research Funds for the Central Universities No.2019kfyXKJC024, and the 111 Project on Computational Intelligence and Intelligent Control under Grant B18024.

	{\small
		\bibliographystyle{iccv21}
		\bibliography{egbib}
	}
	
	\clearpage
	\appendix
	\section{Additional Experimental Results}
	\subsection{Paths for Evaluation}
	In our paper, the instance path is only applied in training, facilitating the identity feature learning of the main search path. During inference, we drop the instance path and the images are only passed through the search path. We compare the results of using different paths for testing, as shown in Tab.~\ref{app:path}. It can be seen that using two paths for evaluation cannot bring extra performance gains. This indicates the context-invariant embeddings produced by our framework.
	\begin{table}[htbp]
		\small
		\begin{center}
			\caption{Comparisons of using different paths for evaluation on the CUHK-SYSU dataset.}
			\vspace{1mm}
			\begin{tabular}{ccc}
				\toprule
				Inference path& mAP& Rank-1  \\
				\midrule
				\midrule
				Two paths & 85.65 & 86.73 \\
				Search path & \textbf{85.72} & \textbf{86.86} \\
				\bottomrule
			\end{tabular}
			\label{app:path}
		\end{center}
	\end{table}
	
	\vspace{-2mm}
	\subsection{Different Detection Networks}
	Following~\cite{han2021decoupled}, we choose the RepPoints as the detection network. To show the expandability of our R-SiamNet, different detectors are incorporated into our framework, including Faster R-CNN~\cite{ren2015faster}, RetinaNet~\cite{lin2017focal} and RepPoints~\cite{yang2019reppoints}. As reported in Tab.~\ref{app:detector}, the final performance gaps among different detectors are small, exhibiting the effectiveness and robustness of our framework.
	
	\begin{table}[htbp]
		\small
		\begin{center}
			\caption{Comparisons when incorporated with different detectors on the CUHK-SYSU dataset.}
			\vspace{1mm}
			\begin{tabular}{ccc}
				\toprule
				Detector &mAP&Rank-1               \\ 
				\midrule
				Faster R-CNN  & 84.84& 85.72      \\
				RetinaNet &85.39&86.59             \\
				RepPoints  & \textbf{85.72} & \textbf{86.86}             \\
				\bottomrule
			\end{tabular}
			\label{app:detector}
		\end{center}
	\end{table}
	\vspace{-2mm}
	\subsection{Comparisons with Two-Step Manner}
	We combine a well-trained RepPoints detector~\cite{yang2019reppoints} and an unsupervised re-ID model called SPCL~\cite{ge2020self} as our two-step competitor. As shown in Tab.~\ref{app:two-step}, our method outperforms it by a large margin with higher efficiency. It shows that training detection and identification end-to-end is beneficial for obtaining better representations. It may also imply the importance of instance-level consistency learning. 
	\begin{table}[htbp]
		\small
		\begin{center}
			\caption{Comparisons with two-step manner on the CUHK-SYSU dataset.}
			\vspace{1mm}
			\begin{tabular}{ccc}
				\toprule
				Methods&mAP&Rank-1               \\ 
				\midrule
				RepPoints+SPCL  & 73.43& 74.79      \\
				Ours &\textbf{85.72} & \textbf{86.86}             \\
				\bottomrule
			\end{tabular}
			\label{app:two-step}
		\end{center}
	\end{table}
	
\subsection{Evaluation on filter strategy in the clustering.} 
To analyze the effectiveness of the filter strategy in clustering, we conduct experiments with/without filtering by image information. As shown in Tab.~\ref{tab:filter}, we observe $1.22\%/1.13\%$ rank-1 drops on CUHK-SYSU/PRW datasets by removing the filter strategy. This shows that it is beneficial to filter the aggregation of the persons from the same scene images.

\begin{table}[htbp]
	\small
	\centering
	\caption{Performance of our method with/without the filter strategy in clustering. Results on the CUHK-SYSU and PRW datasets are shown. R-SiamNet w/o filter means the clustering is applied without filtering by image information.}
	\vspace{1mm}
	\label{tab:filter}
	\begin{tabular}{ccccc}
		\toprule	
		\multirow{2}{*}{Methods} &
		\multicolumn{2}{c}{CUHK-SYSU} &
		\multicolumn{2}{c}{PRW} \\
		\cmidrule{2-5}
		~&mAP&Rank-1&mAP&Rank-1 \\
		\midrule	
		\midrule				
		R-SiamNet w/o filter &84.74&85.64  & 20.31&72.23\\
		R-SiamNet   &\textbf{85.72} &\textbf{86.86} & \textbf{21.16}&\textbf{73.36}\\
		\bottomrule	
	\end{tabular}
\end{table}

	\subsection{Different numbers of training epochs}
\begin{figure}[htbp]
		\small
		\begin{center}
			\includegraphics[width=0.96\linewidth]{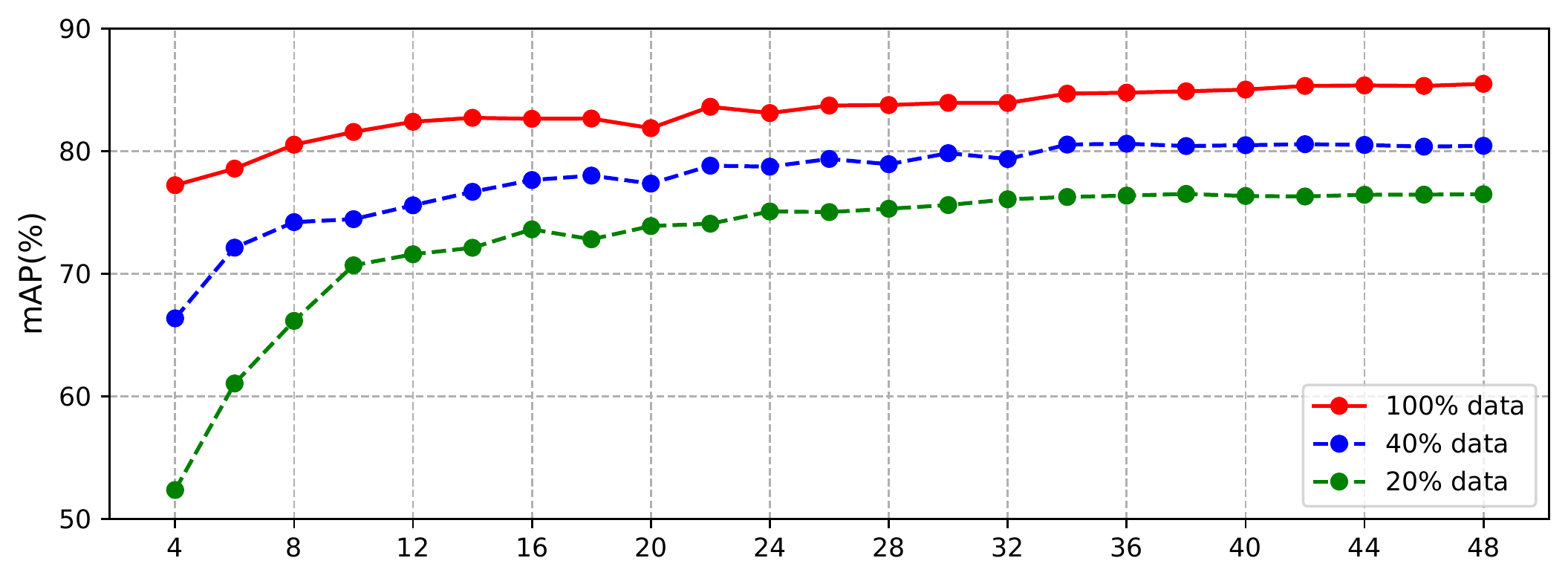} 
		\end{center}
		\caption{Performance with different numbers of training epochs. The results of three data scales are exhibited on the CUHK-SYSU dataset. }
		\label{app:app_datasize}
	\end{figure}

	We illustrate the mAP scores with different numbers of training epochs. As Fig.~\ref{app:app_datasize} shows, the results of three data scales are exhibited on the CUHK-SYSU dataset. It can be observed that the performance improves steadily to saturation as the epoch increases. With smaller data scales, the mAP reaches saturation earlier.
	
	\subsection{Runtime Comparisons} 
	To compare the evaluation efficiency of our framework with other methods, we report the average runtime of the inference stage for a whole scene image, shown in Tab.~\ref{tab:runtime}. Since these methods are evaluated with different GPUs, we also exhibit the Tera-Floating Point Operation per-second (TFLOPs) for fair comparisons. Similar to other methods~\cite{chen2020norm,munjal2019query,chen2018person}, we evaluate the models with an input image size of $900 \times 1500$. As shown in Tab.~\ref{tab:runtime}, our R-SiamNet takes $72$ milliseconds to process one image, which is faster than the two-step method MGTS~\cite{chen2018person} by a large margin. The query-guided method QEEPS~\cite{munjal2019query} requires to re-compute all the gallery embeddings for each query image. This time-consuming operation reduces the evaluation efficiency. Moreover, our method is $13\%$ faster than NAE~\cite{chen2020norm}. These results clearly demonstrate the efficiency of our R-SiamNet in evaluation.
	\begin{table}[htbp]
		\small
		\begin{center}
			\caption{Runtime comparisons of different methods when evaluation. The average runtime for one image with the size of $900 \times 1500$ is exhibited on the CUHK-SYSU dataset. }
			\vspace{2mm}
			\begin{tabular}{ccc}
				\toprule
				Methods& GPU (TFLOPs)& Runtime (ms)  \\
				\midrule
				\midrule
				MGTS~\cite{chen2018person}&K80 (8.7)&1269 \\
				QEEPS~\cite{munjal2019query}&P6000 (12.0)&300 \\
				NAE+~\cite{chen2020norm}&V100 (14.1)&98 \\
				NAE~\cite{chen2020norm}&V100 (14.1)&83 \\
				Ours&V100 (14.1)&\textbf{72}  \\
				\bottomrule
			\end{tabular}
			\label{tab:runtime}
		\end{center}
	\end{table}
	
	\begin{figure*}[!ht]
		
		\centering
		\subfigure[R-SiamNet w/o $L_{ins}$ \& $L_{int}$ \& $L_{clu}$]{%
			\label{app:app_tsne1}%
			\includegraphics[width=0.345\linewidth]{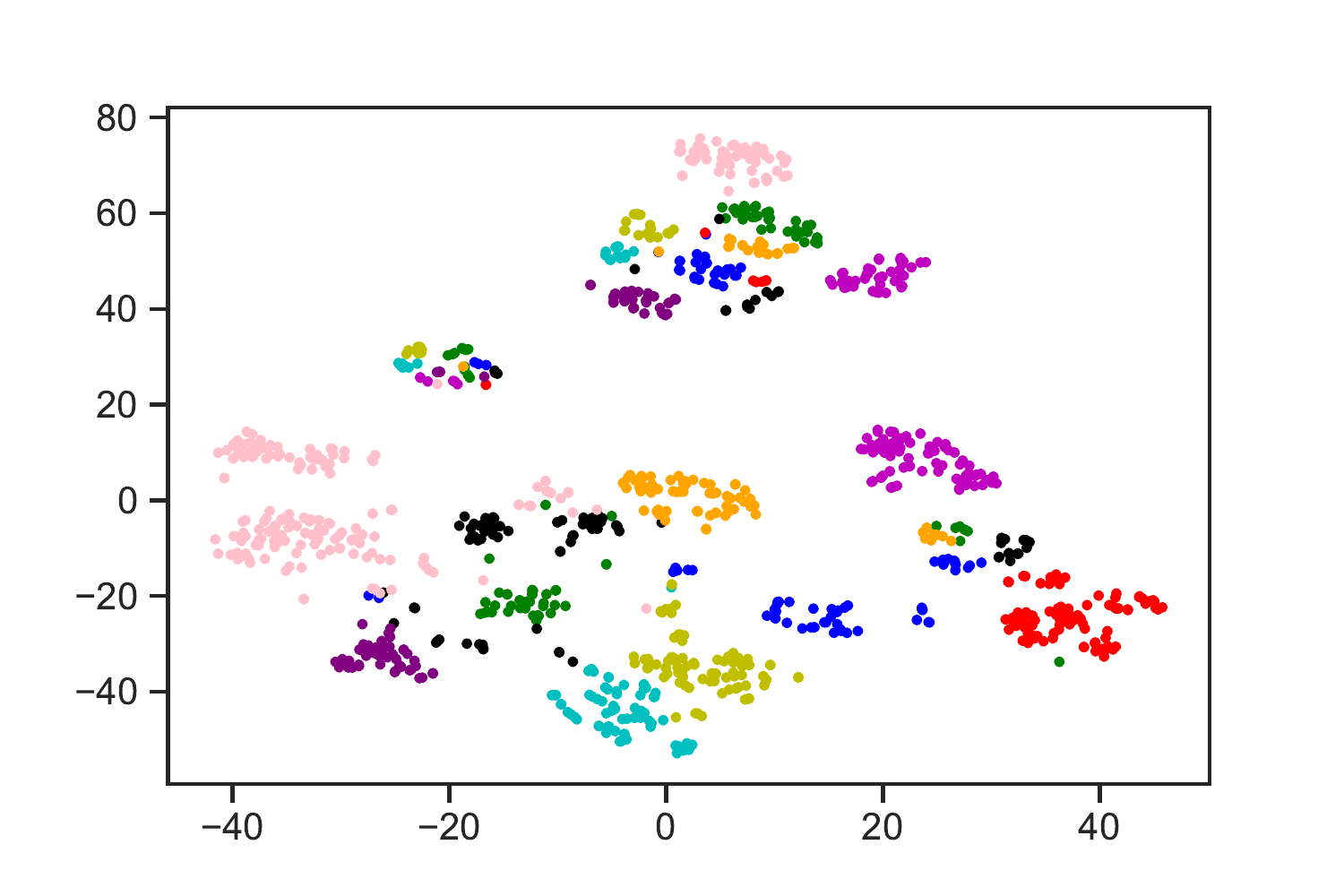}}%
		\subfigure[R-SiamNet w/o $L_{clu}$ ]{%
			\label{app:app_tsne2}%
			\includegraphics[width=0.345\linewidth]{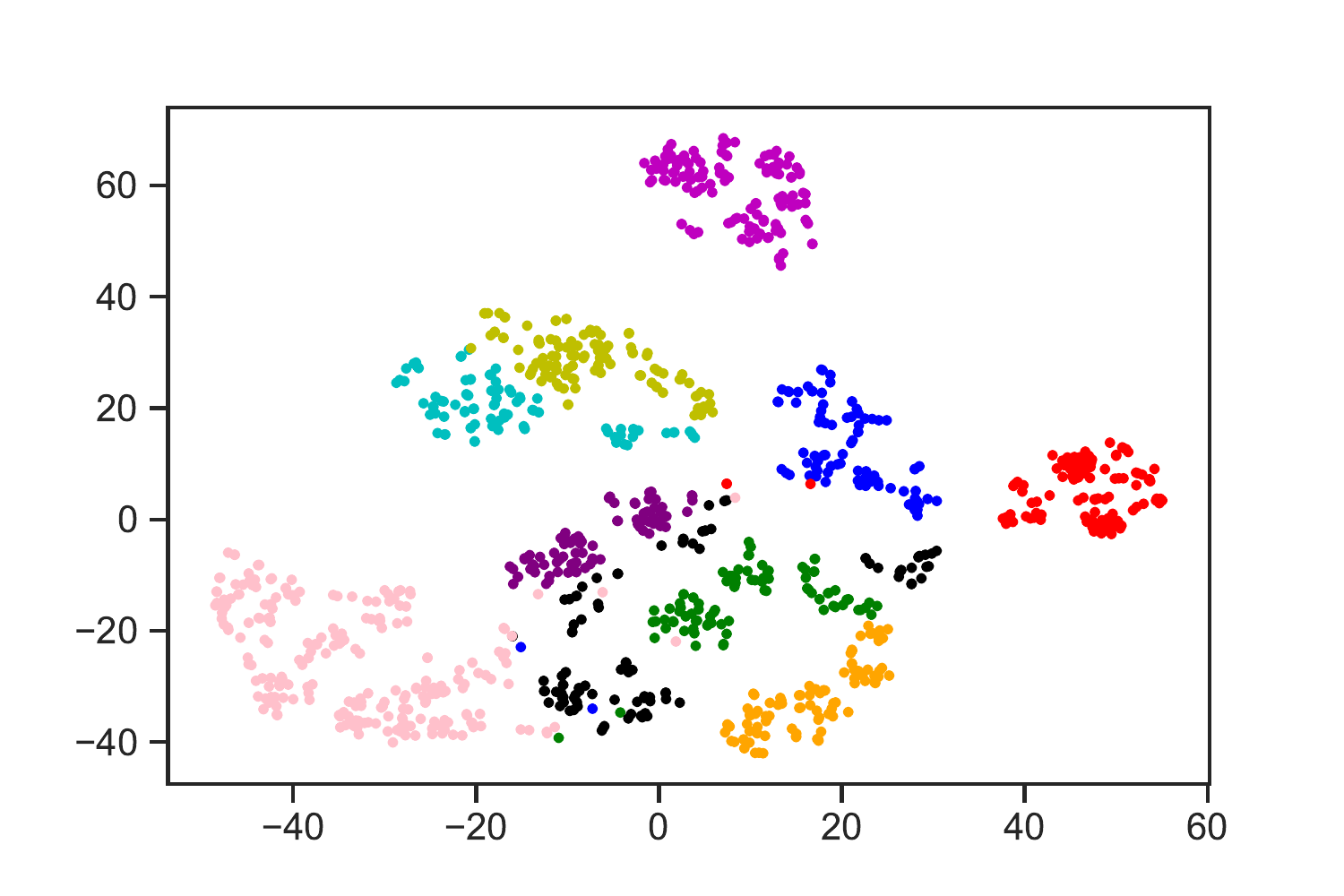}}%
		\subfigure[R-SiamNet]{%
			\label{app:app_tsne3}%
			\includegraphics[width=0.345\linewidth]{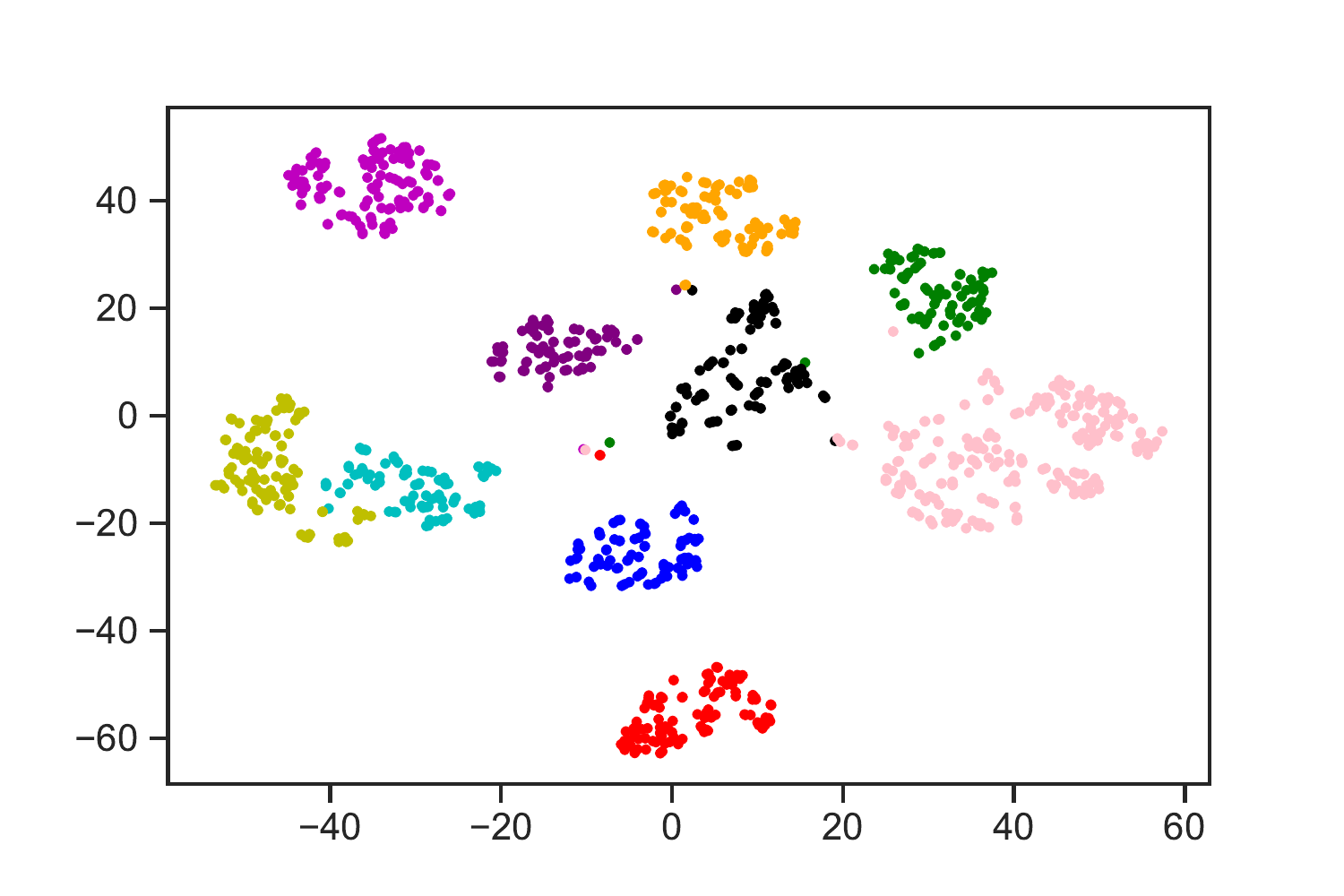}}%
		\vspace{2mm}
		\caption{T-SNE feature visualization on a part of the PRW training set ($10$ classes, $1,089$ pedestrians). (a) R-SiamNet without $L_{ins}$ \& $L_{int}$ \& $L_{clu}$. (b) R-SiamNet without $L_{clu}$. (c) Our proposed R-SiamNet with both instance-level consistency learning and cluster-level contrastive learning. Colors denote person identities.} 
		\label{app:tsne}
	\end{figure*}
	
	\section{More Qualitative Analysis}
	\subsection{Feature Visualization} 
	To analyze the discriminative ability of our learned features, we employ the t-SNE~\cite{van2008visualizing} to visualize the feature representations in training. As illustrated in Fig.~\ref{app:tsne}, there are $1,089$ pedestrians with $10$ classes, which is a subset of the PRW training set. Different colors represent different classes. 
	
	Fig.~\ref{app:app_tsne1} shows the result when training with a single search path. Without applying the instance-level consistency learning and cluster-level contrastive learning, the learned features show large intra-class distances and small inter-class distances. When adding the instance-level consistency learning, including $L_{ins}$ and $L_{int}$, the result is shown in Fig.~\ref{app:app_tsne2}. 
	It is observed that the feature embeddings of the same category can be aggregated compared with Fig.~\ref{app:app_tsne1}. This shows the effectiveness of our instance-level consistency learning. Nevertheless, the features within the class are gathered loosely, and the margins among different classes are not clear. Furthermore, we apply the cluster-level contrastive learning, and the result is shown in Fig.~\ref{app:app_tsne3}. It can be seen that both intra-class compactness and inter-class separability are further encouraged. There are obvious margins among most categories. This verifies that our R-SiamNet can generate discriminative embeddings under the weakly supervised settings.

\end{document}